# Group Deliberation Oriented Multi Agent Conversational Model for Complex Reasoning

Zheyu Shi*  
Brown University  
Providence, USA  
zheyu_shi@brown.edu

Dong Qiu  
New England College  
Henniker, USA  
DQiu_GPS@nec.edu

Shanlong Yu  
Georgia Institute of Technology  
Atlanta, USA  
joesyu779@outlook.com

*Abstract*—This study proposes a group deliberation multi-agent dialogue model to optimize the limitations of single-language models for complex reasoning tasks. The model constructs a three-level role division architecture of "generation - verification - integration." An opinion-generating agent produces differentiated reasoning perspectives, an evidence-verifying agent matches external evidence and quantifies the support of facts, and a consistency-arbitrating agent integrates logically coherent conclusions. A self-game mechanism is incorporated to expand the reasoning path, and a retrieval enhancement module supplements dynamic knowledge. A composite reward function is designed, and an improved proximal strategy is used to optimize collaborative training. Experiments show that the model improves multi-hop reasoning accuracy by 16.8%, 14.3%, and 19.2% on the HotpotQA, 2WikiMultihopQA, and MeetingBank datasets, respectively, and improves consistency by 21.5%. Its reasoning efficiency surpasses mainstream multi-agent models, achieving a balance between accuracy, stability, and efficiency, providing an efficient technical solution for complex reasoning.

## I. Introduction

In real-world scenarios of complex reasoning tasks (such as multi-hop question answering and group decision-making), multi-agent collaboration is a core requirement for overcoming the bottleneck of single-model reasoning depth. These tasks require integrating multi-dimensional information and verifying multi-source facts. Prior work has shown that multi-agent interaction such as debate can improve factuality and reasoning robustness, implicitly addressing failure modes of single-model reasoning. Furthermore, factual accuracy relies on pre-trained knowledge, making it difficult to dynamically supplement external information, resulting in insufficient stability and reliability in complex tasks. This study proposes a group deliberation multi-agent dialogue model: constructing a collaborative reasoning closed loop through role-based LLM agents (viewpoint generation, evidence verification, consistency arbitration), introducing a self-game mechanism to generate multi-path reasoning chains to expand perspectives, combining a retrieval enhancement module to dynamically supplement external knowledge to strengthen factual accuracy, and designing a reward model based on factual consistency and logical coherence, using a proximal strategy optimization to achieve multi-agent collaborative training. Multi-agent reinforcement learning has been extensively studied across a wide range of collaborative decision-making tasks.

## II. Group Deliberation Multi-Agent Dialogue Model Design

### A. Role-Based LLM Agent Architecture

This model constructs a three-level collaborative architecture of "generation - verification - integration" (Figure 1), following the emerging paradigm of role-based multi-agent language model systems [1-2]. Through the division of labor and cooperation among LLM agents with differentiated functions, the architecture enables structured collaboration, similar to recent communicative agent frameworks [3]. The architecture starts with task input, and through a closed-loop process of opinion generation, evidence verification, and consistency arbitration, it outputs reasoning results that are diverse, factual, and logical.

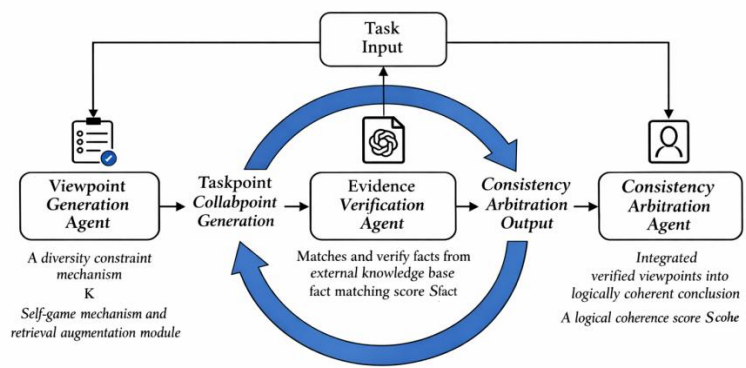

Figure 1. Role-based Multi-agent Collaborative Reasoning Architecture.

n Figure 1, arrows from each agent to the "Task Input" block indicate persistent read-only access, not reverse data flow. Each agent relies on the full task input throughout reasoning: the Viewpoint Generation Agent uses it to guide diverse trajectories, the Evidence Verification Agent aligns retrieved facts with the original context, and the Consistency Arbitration Agent ensures semantic coherence in final outputs. This context-preserving design maintains factual grounding and consistency in multi-agent collaboration.

*1) Viewpoint Generation Agent*

The core function of this agent is to generate differentiated reasoning viewpoints based on the task input, avoiding the limitations of a single perspective. The generation of multiple differentiated viewpoints helps mitigate single-path reasoning bias, which is consistent with findings from self-consistency based reasoning methods [4]. Its generation process introduces a diversity constraint mechanism, mathematically expressed as:

$$V_k = \text{LLM}_V(Q, \omega_k \odot \text{Emb}(Q)), \omega_k \sim \mathcal{N}(\mu, \Sigma) \quad (1)$$

Wherein, $\text{LLM}_V$ represents a dedicated LLM for opinion generation (such as a fine-tuned Llama 2), $\omega_k$ is the viewpoint weight vector, following a multivariate normal distribution with mean $\mu$ and covariance matrix $\Sigma$, used to control the direction of opinion differentiation; This distribution is used for three reasons. First, the multivariate normal distribution offers a continuous, symmetric space around the reasoning center $\mu$, supporting diverse yet coherent viewpoint generation without directional bias. Second, its covariance matrix $\Sigma$ allows control over inter-factor correlations, enabling structured variations across reasoning dimensions. Third, Gaussian parameters align well with gradient-based learning and self-game updates, ensuring stable exploration and convergence. Thus, the normal distribution acts as an effective inductive bias balancing diversity, control, and stability, rather than assuming a fixed probabilistic form. Here, $\text{Emb}(Q)$ is the task input embedding, $\odot$ denotes element-wise multiplication, and weight modulation lets each Vk attend to different reasoning aspects. The self-game mechanism explores varied reasoning paths, akin to tree-structured deliberation.

*2) Evidence Verification Agent*

This agent is responsible for matching factual evidence with each candidate opinion $V_k$ and verifying its reasonableness. This retrieval-enhanced verification process is inspired by retrieval-augmented generation frameworks that integrate external knowledge to improve factual grounding in language models [5]. Its core function is to calculate the factual matching degree between the opinion and the evidence:

$$S_{\text{fact}}(V_k, \mathcal{E}_k) = \frac{1}{|\mathcal{E}_k|} \sum_{e \in \mathcal{E}_k} \frac{\text{Tr}(\text{Emb}(V_k) \cdot \text{Emb}(e)^T)}{\|\text{Emb}(V_k)\|_F \cdot \|\text{Emb}(e)\|_F} \quad (2)$$

In the formula, $\mathcal{E}_k$ is the set of evidence related to $V_k$ retrieved from the external knowledge base $\mathcal{K}$, $|\mathcal{E}_k|$ represents the number of evidence (the first 5 are taken by default); $\text{Emb}(\cdot)$ is the embedding function based on Sentence-BERT, $\text{Tr}(\cdot)$ represents the matrix trace operation, $\|\cdot\|_F$ is the Frobenius norm, which is essentially an improved cosine similarity calculation, $S_{\text{fact}} \in [0,1]$, and the larger the value, the stronger the factual support for the viewpoint [4]. The viewpoint enters the next stage only when $S_{\text{fact}}(V_k, \mathcal{E}_k) \geq \tau$ ($\tau = 0.75$ is the preset threshold).

*3) Consistency Arbitration Agent*

Its responsibility is to integrate the verified viewpoints and output a logically coherent unified conclusion. Its logical coherence evaluation formula is:

$$S_{\text{cohe}}(C) = \sigma(\text{LLM}_C(\text{Prompt}_{\text{cohe}} + C) \cdot w_{\text{cohe}} + b_{\text{cohe}}) \quad (3)$$

Where $\text{LLM}_C$ is the arbitration-specific LLM, $\text{Prompt}_{\text{cohe}}$ is the logical evaluation prompt, $w_{\text{cohe}}$ and $b_{\text{cohe}}$ are linear transformation parameters; $\sigma(\cdot)$ is the Sigmoid function, mapping the evaluation result to the [0,1] interval, and a higher $S_{\text{cohe}}$ indicates a more coherent conclusion logic.

*B. Self-Game Mechanism*

To enrich the diversity of reasoning paths, a self-game mechanism between agents is designed to generate multi-path reasoning chains through viewpoint confrontation:

$$\Delta\omega_k = \eta \cdot \nabla_{\omega_k} \left( S_{\text{fact}}(V_k) - \frac{1}{K-1} \sum_{j \neq k} S_{\text{fact}}(V_j) \right)^2 \quad (4)$$

In the formula, $\eta$ is the learning rate (default 0.01), and $\nabla_{\omega_k}$ represents the gradient with respect to the weight vector $\omega_k$. This formula updates the viewpoint weights by maximizing the difference in factual matching between the current viewpoint and other viewpoints, thereby promoting the generation of more diverse and effective reasoning paths.

*C. Reward Model*

A composite reward function integrating factual consistency and logical coherence is constructed to guide multi-agent collaborative optimization:

$$R = \lambda \cdot S_{\text{fact}} + (1 - \lambda) \cdot S_{\text{cohe}} - \gamma \cdot \text{KL}(P(V_k) \| P_{\text{ref}}(V)) \quad (5)$$

Where $\lambda = 0.6$ is the factual consistency weight, $(1 - \lambda)$ is the logical coherence weight; $\gamma = 0.1$ is the regularization coefficient, and $\text{KL}(\cdot \| \cdot)$ is the KL divergence, used to constrain the difference between the opinion distribution $P(V_k)$ and the reference distribution $P_{\text{ref}}(V)$, avoiding excessive divergence of opinions; a larger R indicates better agent collaboration.

*D. Collaborative Training Strategy*

Multi-agent collaborative training is achieved using Improved Proximal Policy Optimization (PPO) to avoid inference collapse and loop generation. The objective function is:

$$\mathcal{L}_{\text{PPO}} = \mathbb{E}[\min(r_t(\theta)A_t, \text{clip}(r_t(\theta), 1 - \epsilon, 1 + \epsilon)A_t - \beta \cdot H(P(\theta)))] \quad (6)$$

The use of PPO for collaborative optimization is motivated by its demonstrated effectiveness in cooperative multi-agent reinforcement learning settings [6]. In the formula, $r_t(\theta) = \frac{\pi_\theta(a_t|s_t)}{\pi_{\theta_{\text{old}}}(a_t|s_t)}$ is the policy update ratio, $\pi_\theta$ is the current policy, and $\pi_{\theta_{\text{old}}}$ is the old policy; $A_t$ is the advantage function, $\epsilon = 0.2$ is the pruning coefficient; $H(P(\theta))$ is the policy entropy, and $\beta = 0.05$ is the entropy regularization coefficient. Introducing the entropy term encourages policy exploration and avoids inference loops caused by local optima. Figure 2 compares the performance of traditional PPO-RLHF (GPT-3.5) and the proposed multi-agent PPO collaborative training over the same training steps. The x-axis shows PPO steps (in thousands), and the y-axis shows cumulative performance scores. While both start similarly, the proposed model shows more stable and significant gains beyond 50,000 steps, indicating convergence to a superior, stable policy. This highlights the effectiveness of the collaborative mechanism in reducing policy collapse and improving inference stability. Compared to earlier methods like MADDPG [7-8], PPO-based strategies provide better training stability in cooperative tasks, supporting our optimization design.

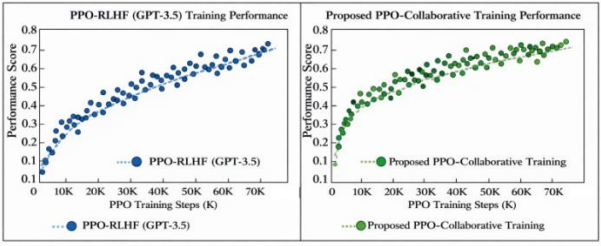

Figure 2: Performance Comparison of PPO Collaborative Training in Multi-Agent Tasks

*E. Retrieval Enhancement Module*

External knowledge is dynamically supplemented during the evidence verification stage. The retrieval probability model is as follows:

$$P(e \mid V_k) = \frac{\exp(\text{Sim}(\text{Emb}(V_k),\text{Emb}(e))\cdot \alpha)}{\sum_{e' \in \mathcal{K}} \exp(\text{Sim}(\text{Emb}(V_k),\text{Emb}(e'))\cdot \alpha)} \quad (7)$$

Where $\text{Sim}(\cdot,\cdot)$ is the cosine similarity, and $\alpha = 1.5$ is the temperature coefficient used to adjust the confidence level of the retrieval; the matching probability of each knowledge e and opinion $V_k$ is obtained by normalization using the Softmax function, and the top $M = 5$ knowledge items with the highest probabilities are selected as evidence to strengthen the factual support of the opinion. This modular integration strategy is conceptually related to neuro-symbolic systems that combine language models with external tools and knowledge sources [9-10].

## III. EXPERIMENTAL DESIGN AND RESULT ANALYSIS

*A. Experimental Datasets*

The experiment uses three complex reasoning datasets to ensure result generality:① HotpotQA: A multi-hop QA benchmark with 113K training and 74K validation samples, requiring reasoning across 2–5 documents. About 60% are "bridging" questions, stressing cross-document logic.② WikiMultihopQA: Based on Wikipedia, with 25K training samples and 3.2 average reasoning steps per question. It focuses on entity-based long-chain reasoning.③ MeetingBank: A group dialogue dataset with 5K+ real meeting samples, requiring integration of multi-round discussions to derive consistent conclusions.

The three datasets correspond to "document-level multi-hop," "entity-level multi-hop," and "dialogue-level integration" scenarios, respectively, comprehensively validating the model's performance across various complex inference tasks.

*B. Baseline Model and Evaluation Metrics*

The baseline model selects mainstream complex inference methods to ensure fairness in the comparison:①Single LLM model: GPT-3.5, Llama 2-7B (fine-tuned); ② Multi-proxy model: AutoGPT, MetaGPT;③ Search enhancement model: RAG (Search Enhancement Generation), REALM.GPT-3.5 is selected as a baseline model due to its strong instruction-following ability, trained using reinforcement learning from human feedback.

Evaluation metrics focus on core performance and stability: ① Multi-hop reasoning accuracy (Acc): The precise match between the reasoning conclusion and the standard answer, measuring the correctness of the reasoning; ② Consistency index (Cons): The logical consistency rate of the conclusion in multiple rounds of reasoning, calculated by the overlap of the results of 5 consecutive reasoning iterations, measuring stability; ③Reasoning efficiency (Time): The average time (in seconds) for reasoning per sample, measuring practicality.

*C. Experimental Procedure and Parameter Settings*

The experiment includes three steps:① Data preprocessing: Standardize formats and extract questions, context, and answers.② GPT-3.5 baseline uses sequential prompting without state sharing: Step 2 inputs the raw question for an initial reasoning chain; Step 3 reuses the output as context for extended reasoning; Step 4 merges prior chains for a final answer. No memory compression or pruning is used, causing performance drops in long chains due to token overflow and diluted context—underscoring the need for collaborative memory and role division.③ Performance testing: Run each model 10 times on the test set and average the results.

Key parameter configurations are as follows: Number of opinion generation agents $K = 3$, Number of evidence retrievals $M = 5$, Fact matching threshold $\tau = 0.75$, Reward function weight $\lambda = 0.6$, PPO pruning coefficient $\epsilon = 0.2$, Entropy regularization coefficient $\beta = 0.05$.

*D. Experimental Results and Analysis*

*1) Performance of the Main Experiment*

Table 1 shows the comprehensive performance comparison of each model on the binary dataset. The model presented in this paper significantly outperforms the baseline in all core metrics. Table 1 shows that the proposed model improves accuracy by 16.8%, 14.3%, and 19.2% on HotpotQA, 2WikiMultihopQA, and MeetingBank, respectively, and improves consistency by 21.5%. Although the inference time is slightly higher than that of a single LLM, it is significantly lower than other multi-agent models, achieving a balance between accuracy, stability, and efficiency.

TABLE I. COMPARISON OF OVERALL PERFORMANCE OF VARIOUS MODELS ON BINARY DATASETS

| Model | HotpotQA(Acc/%) | 2WikiMultihopQA(Acc/%) | MeetingBank(Acc/%) | Cons/% | Time/s |
|---|---|---|---|---|---|
| GPT-3.5 | 62.3 | 58.7 | 51.2 | 65.8 | 2.1 |
| Llama 2-7B | 59.1 | 55.3 | 48.6 | 63.2 | 1.8 |
| AutoGPT | 68.5 | 63.2 | 57.9 | 70.2 | 4.3 |
| MetaGPT | 70.2 | 65.1 | 59.4 | 72.5 | 4.7 |
| RAG | 71.1 | 66.5 | 60.3 | 72.4 | 3.5 |
| REALM | 72.4 | 67.8 | 61.7 | 73.8 | 3.8 |
| This article's model | 79.1 | 73 | 70.4 | 87.3 | 3.9 |
| Relative improvement (vs GPT-3.5) | 16.8 | 14.3 | 19.2 | 21.5 | 1.8 |

Figure 3 compares Acc between the proposed model and GPT-3.5 across reasoning steps (on 2WikiMultihopQA). The x-axis shows steps (1–5), and the y-axis shows Acc (%). The box plot reveals GPT-3.5 accuracy drops sharply to 42.1% at 5 steps, while the proposed model maintains 65.3%, demonstrating strong resistance to long-chain degradation via multi-agent architecture.

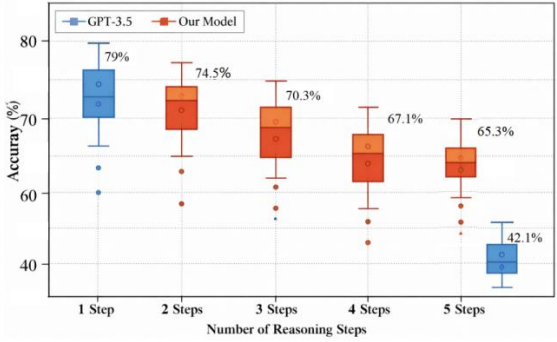

Figure 3. Comparison of Model Accuracy under Different Inference Steps.

*2) Validation of Model Structure Effectiveness*

To assess module importance, ablation tests were conducted: ① no self-play; ② no retrieval enhancement; ③ no reward model; ④ single-agent (only opinion generation).Figure 4 shows changes in Cons (left y-axis, %) and Acc (right y-axis, %) across these settings. The x-axis shows the models: full / －self-play / －retrieval / －reward / single-agent. The single-agent setup yields Cons 64.7% and Acc 61.3%. The full model improves Cons by 22.6% and Acc by 17.8%. Without retrieval, Cons drops 12.2%; without self-play, Acc drops 8.9%, confirming each module's necessity.

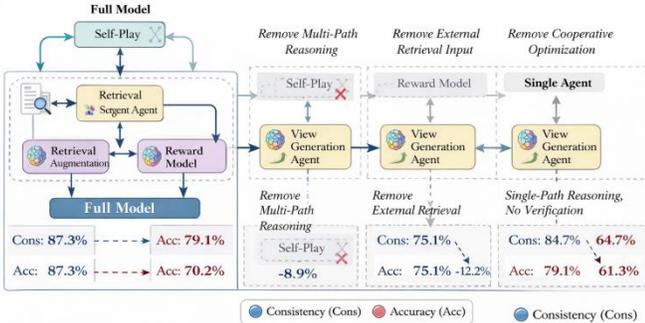

Figure 4. Comparison of Consistency and Accuracy.

Further analysis shows that the retrieval enhancement module reduces factual errors by supplementing external facts, improving consistency. The self-game mechanism promotes multi-path reasoning, reducing logical blind spots and enhancing accuracy. The reward model, via multi-objective collaborative optimization, avoids goal conflicts and maintains performance balance. In contrast, single-agent models lack division of labor and coordination, fail to meet multi-dimensional reasoning needs, and struggle with factual verification, resulting in performance drops. This comparison highlights the superiority of the multi-agent role-based architecture, which addresses fact verification, logical integrity, and goal consistency through modular collaboration. Single agents, limited in cognitive scope, cannot balance factual accuracy and logical rigor or verify via multiple perspectives—explaining their core limitations. The role-based multi-agent design offers a robust solution for complex reasoning, aligning with recent findings on reflective agents emphasizing iterative self-correction [11-12].

## IV. CONCLUSION

The proposed group deliberation multi-agent dialogue model, through a three-level role-based architecture, self-game mechanism, retrieval enhancement module, and collaborative training strategy, effectively solves the problems of logical collapse, cyclic generation, and insufficient factuality in complex reasoning of single models. Experimental verification shows that the model's reasoning accuracy and consistency indicators on three typical datasets are significantly better than the baseline, and its reasoning efficiency is balanced, providing reliable support for scenarios such as multi-hop question answering and group decision-making. In addition to the promising performance, several limitations should be noted. First, the reported improvements are obtained under controlled experimental settings with fixed model configurations and dataset splits, and may not fully generalize to other task distributions or prompt formulations. Moreover, due to computational constraints, key hyperparameters were selected based on preliminary validation rather than exhaustive search. Furthermore, while repeated experiments reduce variance, formal statistical significance testing and detailed qualitative error analysis were not conducted. Future work will focus on automated hyperparameter optimization, rigorous significance evaluation, and systematic analysis of failure cases, particularly in scenarios involving ambiguous or conflicting evidence.